%% file: main_v2.tex
\documentclass[12pt,letterpaper]{article}

\include{Preamble}

\title{Representing affect information in word embeddings}

\author{Yuhan Zhang, Wenqi Chen, Ruihan Zhang \& Xiajie Zhang\footnote{Authors: Yuhan Zhang, Harvard University (\href{mailto:yuz551@g.harvard.edu}{yuz551@g.harvard.edu}); Wenqi Chen, Harvard University (\href{mailto:wenqichen@g.harvard.edu}{wenqichen@g.harvard.edu}); Ruihan Zhang, Massachusetts Institute of Technology (\href{mailto:ruihanz@mit.edu}{ruihanz@mit.edu}); Xiajie Zhang, Massachusetts Institute of Technology (\href{mailto:xiajie@mit.edu}{xiajie@mit.edu}). We would like to thank Mycal Tucker, Roger Levy, and the audience at ELM 2022 for their generous feedback. All mistakes are ours.}}

\begin{document}

\setlength{\Extopsep}{6pt}
\setlength{\Exlabelsep}{6.75pt}		

\maketitle
\thispagestyle{empty}
\begin{abstract}
A growing body of research in natural language processing (NLP) and natural language understanding (NLU) is investigating human-like knowledge learned or encoded in the word embeddings from large language models. This is a step towards understanding what knowledge language models capture that resembles human understanding of language and communication. Here, we investigated whether and how the affect meaning of a word (i.e., \emph{valence}, \emph{arousal}, \emph{dominance}) is encoded in word embeddings pre-trained in large neural networks. We used the human-labeled dataset \citep{mohammad-2018-obtaining} as the ground truth and performed various correlational and classification tests on four types of word embeddings. The embeddings varied in being static or contextualized, and how much affect specific information was prioritized during the pre-training and fine-tuning phase. Our analyses show that word embedding from the vanilla BERT model \citep{devlin-etal-2019-bert} did not saliently encode the affect information of English words. Only when the BERT model was fine-tuned on emotion related tasks or contained extra contextualized information from emotion-rich contexts could the corresponding embedding encode more relevant affect information.
\end{abstract}


\vspace{6pt} 

\section{Introduction}
With the success of large neural network models in completing complicated language tasks, evaluating the models' interpretability and intrinsic capabilities has become a heated research trend \citep[e.g.,][]{manning2020emergent, mikolov2013distributed}. The evaluation work could be roughly classified into two types: one that relies on the output of the language models (LMs) to infer the model's linguistic ability and the other that looks into the components of LMs (e.g., word embeddings) for such inspiration. While previous evaluation tasks have focused on testing LMs' explicit linguistic knowledge (e.g., syntactic knowledge such as islands, semantic knowledge such as compositionality, word-level knowledge such as polysemy), we pick a piece of knowledge that is less studied but essential to intelligence. Specifically, we studied whether and how word embeddings learned via supervised methods in large neural networks encode the affect information of a word (e.g., valence, arousal, dominance). Our work shows that even though contextualized word embeddings were in general better at capturing intricate affect meanings compared to static word embeddings, especially after being fine-tuned on emotion related tasks, word embeddings from vanilla BERT did not attain salient affect knowledge.

In Section \ref{related-work}, we detailed relevant work that led us to our investigation. In Section \ref{method}, we laid out the unsupervised and supervised methodologies we took and in Section \ref{results}, we spelled out the findings. In Section \ref{discussion}, we discussed implications of our research and possible future directions.


\section{Related work} \label{related-work}

\subsection{What can language models learn?}

A growing body of research has investigated what linguistic knowledge large artificial neural networks can learn while they are trained in a supervised way to predict the next word. In the syntactic category, scholars have shown that large LMs have varying grammatical knowledge ranging from the local subject-verb agreement to long-distance filler-gap dependencies \citep[e.g.,][]{hu-etal-2020-systematic, linzen2016assessing, warstadt2020blimp, wilcox-etal-2018-rnn}. Aside from looking at predicted words to infer LMs' linguistic knowledge, \cite{manning2020emergent} show that a linear transformation of word embeddings from BERT \citep{devlin-etal-2019-bert} captures linguistic hierarchical structures. There is also increasing attention to understanding LMs' abilities to represent meanings. In semantics and pragmatics, promising and positive results seem to support LMs' increasingly sophisticated abilities such as doing natural language inference \citep[e.g.,][]{poliak-etal-2018-collecting-diverse, wang-etal-2018-glue}. Instead of relying on direct output of LMs, finding the relationship between human understanding of language and representations in word embeddings have also been fruitful.

\subsection{Word embedding and lexical semantics}
Since deep contextual language models provide contextualized word embeddings which naturally encode the distance between word tokens in a vector space, this property can be utilized to test whether the trained distance in word embeddings reflects the natural way words group together according to our lexical semantic knowledge. Existing studies have shown that the pre-trained BERT model is able to place polysemous words that appear in different contexts into distinct regions of the shared vector space \citep{wiedemann-2019-bert} and the word sense distances are correlated with human judgments \citep{nair-etal-2020}. There is also the general claim that contextualized word embedding in BERT is good at word sense disambiguation \citep{loureiro-2021}. But in addition to this line of research on word sense disambiguation, investigation into other aspects of lexical semantics is limited. We made an attempt in this work to bring together studies of word embedding representations and another aspect of lexically encoded meaning -- the affect information -- as a novel case study of what LMs can learn from supervised training.

\subsection{Affect information in nlp} 
Affect information refers to any explicit or implicit emotion related information. We use the word \emph{affect} instead of \emph{emotion} because we do not rely on common emotion words such as \emph{happy} and \emph{sad} as baselines of comparison; rather, we rely on three primary independent dimensions of emotions as scales to quantify the meaning of emotion. The three dimensions are \emph{valence} (positiveness-negativeness/pleasantness-unpleasantness), \emph{arousal} (active-passive, some people also take it to mean the intensity of the emotion invoked by the word), and \emph{dominance} (dominant-submissive, or the level of control exerted by the word) \citep{osgood1957measurement, russell1980circumplex, russell2003core}. Another reason to choose this terminology is due to its validity and consistency: there are already multiple human-labeled datasets that are based on these scales as the quantitative ground truth \citep{bradley1999affective, mohammad-2018-obtaining, warriner2013norms}. 

In fields related to emotion recognition, sentiment analysis, and affective computing, transformer-based models and more fine-grained annotation of emotion categories have gathered the attention of the NLP/NLU community \citep[e.g.,][]{alswaidan2020survey, demszky-etal-2020-goemotions,rashkin2018towards, suresh2021using}. Zooming into existing studies relevant to word embeddings and affect information, it has been shown that multi-dimensional embeddings of emotional words achieved higher performance in sentiment analysis tasks than human annotated emotion vectors \citep{li-affective-2017}. Furthermore, the multi-dimensional word embeddings for emotional words are often added onto base static word embeddings for downstream sentiment analysis tasks \citep{mao-2019-sentiment}. While existing research focused more on the application of emotional word embeddings onto sentiment analysis tasks, seldom has explored whether state-of-the-art contextualized word embeddings have learned the affect meaning of all kinds of words. Our study is set up to fill the gap.

\section{Methodology} \label{method}
\subsection{Affect ratings of human as the ground truth}
We selected the NRC VAD dataset (National Research Councial (Canada) Valence Arousal Dominance) curated by \cite{mohammad-2018-obtaining} as the ground truth for affect meanings of English words \footnote{Data can be downloaded from \href{http://saifmohammad.com/WebPages/nrc-vad.html}{http://saifmohammad.com/WebPages/nrc-vad.html}}. There are 20,000 English words annotated along the three independent dimensions of affect. In each dimension, the rating starts from 0 to 1 where 0 means the most negative, passive, or submissive in the V, A, or D dimension and 1 means the most positive, active, or dominant in their respective dimension. This is the largest manually created VAD corpus in any language.   

\subsection{Word embeddings under study} \label{intro-word-embedding}

In this paper, we studied four kinds of word embeddings as described below. We took from each word embedding the same subset of words that are compatible with the NRC VAD corpus. The embeddings differ from each other by how much and what kinds of contextualized information are captured from training. We took GloVe as a representative of existing static word embeddings in the NLP community and compared it with more contextualized ones. We also varied the kinds of contextualized information by prioritizing emotion related and sentiment analysis related information.
\begin{itemize}
    \item Pre-trained word embeddings using the \textbf{GloVe} algorithm \citep{pennington2014glove} which we refer to as the GloVe embedding in this paper. The word embeddings were downloaded from Stanford NLP website\footnote{\href{https://nlp.stanford.edu/projects/glove/}{https://nlp.stanford.edu/projects/glove/}} which were trained on Wikipedia and Gigaword 5 data.
    \item Pre-trained word embeddings that were retrieved from the last hidden layer of the base BERT model \citep{devlin-etal-2019-bert} after we directly ran the model over the NRC word list. We refer to this type of word embedding representation as \textbf{base BERT} throughout the paper.
    \item Pre-trained word embeddings retrieved from the last hidden layer of the BERT model fined tuned on the GoEmotion dataset \citep{demszky-etal-2020-goemotions}. The GoEmotion dataset contains 58k English Reddit comments labeled by humans on 27 emotion categories. We refer to this as \textbf{GoEmotion BERT} in the paper.
    \item Given that the BERT-based word embeddings are supposed to be contextualized because they could be functions of the entire input sentence \citep{ethayarajh-2019-contextual}, we utilized this contextual sensitivity feature and output \textbf{contextualized BERT} word embeddings by deriving them from running IMDB movie reviews \citep{maas-EtAl:2011:ACL-HLT2011}. For each NRC word, we first extracted all its embeddings by retrieving the last hidden layer of the BERT model, we then computed the first principal component to derive the single word vector for each individual word. 
\end{itemize}

\subsection{Unsupervised Probe: Principal Component Analysis}
Firstly, in order to investigate whether the types of word embeddings under investigation encode affect information, we conducted dimensionality reduction via Principal Component Analysis (PCA from the \emph{sklearn} package \citep{sklearn_api, scikit-learn}) over each of the studied word embeddings. PCA is a convenient tool for such visualization without us going deep into deciphering the architecture of multi-dimensional word embeddings. Specifically, we sought for the correlation between each principal component from the embeddings and the human ratings of each VAD dimension. Higher correlation indicates that the high-dimensional information in that word embedding is more likely to saliently encode affect information. This unsupervised approach provides a rudimentary insight into whether affect meanings are well captured by the word embeddings being investigated.

\subsection{Unsupervised Probe: Cosine distance for semantic similarity}
It has been shown that cosine distances between vector representations of words indicate the words' semantic relatedness \citep{dumais1988using, mikolov2013distributed, bojanowski2017enriching}. The literature we learned from is \cite{nair-etal-2020}. They investigated whether contextualized word embeddings -- BERT embeddings in this case -- capture human-like distinctions between English word senses, such as polysemy (chicken as animal vs. chicken as meat) and homonymy (bat as mammal vs. bat as sports equipment). For each pair of target word senses, they measured (1) the cosine distance of the two embeddings and (2) the human judgment of the relatedness of the word senses in a 2-dimensional spatial arrangement task. They then applied Spearman rank correlation analysis to these two measurements. The results turn out that the distance of word senses in BERT's embedding space correlated with human judgments and that the correlation of homonyms was higher.

Learning from \cite{nair-etal-2020}, we took a small sample of affect words and ran correlation tests to see the relationship between the word embedding space and the vector space of human judgments (i.e., the VAD 3-dimensional space). Similar to the PCA results, a stronger correlation means a more salient encoding of real-world affect information in the investigated word embeddings. More about the sampling method: \cite{shaver1987emotion} defined six basic emotion categories (i.e., love, surprise, sadness, anger, joy, fear) and a list of affect words in each category, totalling 132 affect words. We took 80 affect words from \cite{shaver1987emotion} (e.g., \textit{disgust}, \textit{envy}, \textit{enjoyment}, \textit{desire}) and calculated the cosine similarity between each and another word, resulting in $80 \times 79$ similarity scores. We did the same thing by iterating over the kind of vector space, from the NRC VAD 3-dimensional human judgment space, to the four kinds of word embeddings introduced in section \ref{intro-word-embedding}, resulting in $80 \times 79 \times 5$ pairwise similarity scores. Then, we did Spearman rank correlation analysis between one and another embedding representations. We adopted a nonparametric correlation like \cite{nair-etal-2020} because, first, the numerical distribution of a given type of word embedding does not satisfy the normality assumption and, second, we cared more about whether the two embedding representations share a monotonic relationship, instead of a strictly linear relationship. 

\subsection{Supervised Probe: Linear Classifier}

In addition, we experimented with supervised learning methods to probe whether the word embeddings can be linearly separable into the VAD dimensions. We trained a one-layer neural network (i.e., a logistic regression model) to predict the binary VAD labels\footnote{We transformed the numerical human rating (0-1) to a binary categorical variable with the threshold of 0.5.} with the embeddings as inputs. The linear classifier was designed to be as simple as possible to reveal information encoded in the embeddings. If the prediction results match well with the human criterion, it is reasonable to conclude that the word embeddings have the capability to represent the affective information.


\section{Analyses \& Results} \label{results}
\subsection{Principal component analysis}
\begin{figure*}[ht]
\includegraphics[width=16cm]{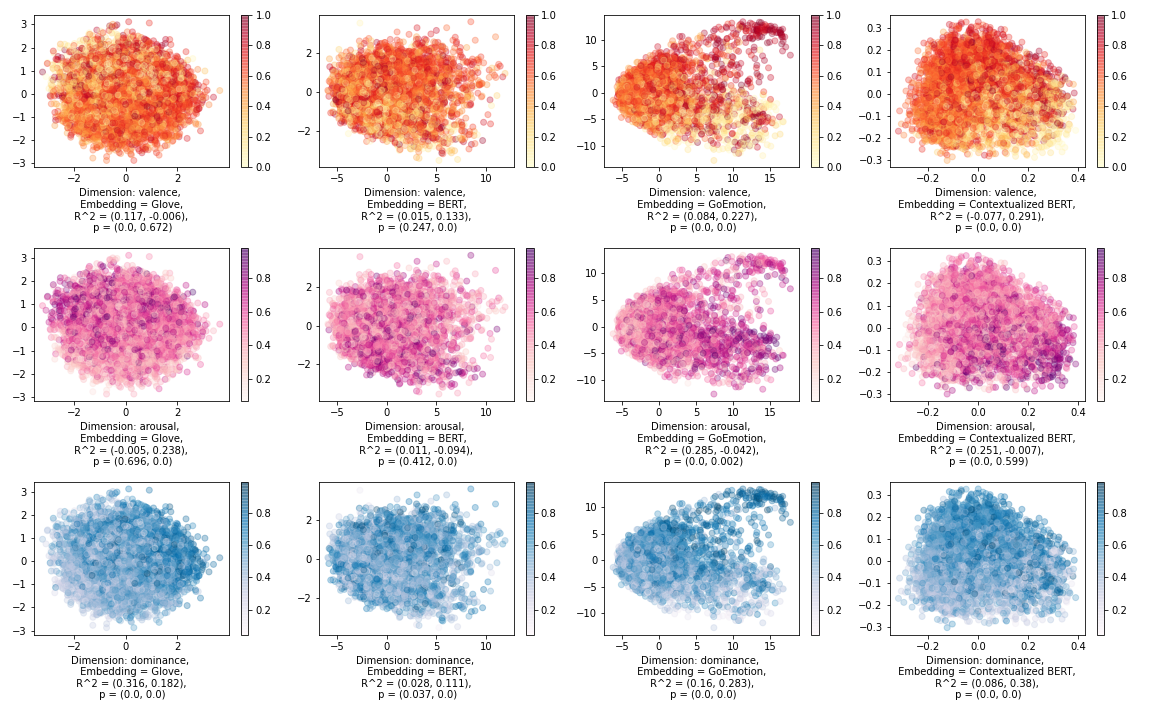}
\caption{2-components PCA representations of GloVe, base BERT, GoEmotion BERT, and contextualized BERT \& human ratings for valence, arousal, and dominance (Human ratings were color-coded in a spectrum. Each dot represents a word. 5,586 words are represented. The darker the dot, the more prominent the human rating in the respective dimension.) }
\label{g:pca_4em}
\end{figure*}

Fig.\ref{g:pca_4em} shows the PCA result. We extracted 5,586 words that are in the vocabulary of GloVe, base BERT, GoEmotion BERT, and contextualized BERT embeddings, ran PCA over these word embeddings, and represented the numerical value of the first two principal components along the two axes. Each horizontal panel represents one dimension of VAD. Each column in Fig.\ref{g:pca_4em} represents a type of tested word embedding. Each point is color coded with human judgment. The darker the color, the higher the rating is along one of the VAD dimensions. Table \ref{pca-corr} shows the Spearman correlation coefficient between the numerical values of human judgments along each affect dimension and the numerical values of the corresponding principal component given a type of word embedding.

From Table \ref{pca-corr}, we learn that every type of word embedding can capture some aspects of the 3-dimensional affect meaning based on the significance level of the correlation coefficients. Yet judging from the magnitude of correlation coefficients for both principal components as well as the number of correlation coefficients that are larger than 0.2 out of the three VAD dimensions, we see that the GoEmotion BERT embedding and the contextualized BERT embedding outperform the other two. 

\begin{table*}[ht]
\centering
\begin{tabular}{c|c|c|c|c|c}
\hline
Dimension & PCA & GloVe & base BERT & GoEmotion BERT & contextualized BERT \\
\hline
\multirow{2}{3em}{Valence} & 1st & 0.117 (.000) & 0.015 (.247) & 0.084 (.000) & -0.077 (.000) \\
& 2nd & -0.006 (.672) & 0.133 (.000) & \textbf{0.227} (.000) & \textbf{0.291} (.000)\\
\hline
\multirow{2}{3em}{Arousal} & 1st & -0.005 (.696) & 0.011 (.412) & \textbf{0.285} (.000) & \textbf{0.251} (.000) \\
& 2nd & \textbf{0.238} (.000) & -0.094 (.000) & -0.042 (.002) & -0.007 (.599)\\
\hline
\multirow{2}{4.2em}{Dominance} & 1st & \textbf{0.316} (.000) & 0.028 (.037) & 0.160 (.000) & 0.086 (.000) \\
& 2nd & 0.182 (.000) & 0.111 (.000) & \textbf{0.283} (.000) & \textbf{0.380} (.000)\\
\hline
\end{tabular}
\caption{Spearman correlation coefficients (\textit{p} values in the bracket) between human ratings (0-1) in each affect dimension and each corresponding principal component of four types of word embeddings (correlation coefficients in bold when larger than 0.2, also see each subplot in Fig.\ref{g:pca_4em})}
\label{pca-corr}
\end{table*}

In Table \ref{explained_variance}, we present the explained variance ratios to show what percentage of the variance in the whole embedding space was explained by the first two principal components for each type of word embedding. Two pieces of observations are worth interpreting. First, the relatively low ratio of explained variance across the eight values indicates that the information represented in the first two principal components might not be representative to generalize on the capacity of certain word embedding to encode certain affect meaning. There might be some affect meaning that has been captured in higher dimensions of the vector space but not projected in the PCA measure. We would need more sensitive probes other than PCA in future investigations. Second, assuming that the PCA analysis is a great approximant for affect encoding, the fact that the first component of base BERT accounts for 29.4\% of the total variance and yet does not correlate with VAD as much as the other word embeddings shows that base BERT might not be optimized to distinguish intricate affect meaning of words. This offers a nice window to peek into what linguistic information is and isn't weighted the most for BERT. 

\begin{table*}[ht]
\centering
\begin{tabular}{c|c|c|c|c}
\hline
Explained ratio & GloVe & base BERT & GoEmotion BERT & contextualized BERT \\
\hline
First & 0.040 & 0.294 & 0.118 & 0.021 \\
\hline
Second & 0.030 & 0.040 & 0.070 & 0.016 \\
\hline
\end{tabular}
\caption{Explained variance ratio of the two principal components}
\label{explained_variance}
\end{table*}

\subsection{Cosine similarity}

Table \ref{correlation} captures the Spearman correlation coefficients based on pairwise cosine similarity of 80 words whose embeddings were from each type of word embeddings. The main comparison is in the first column where human judgments on the semantic similarity of the $80 \times 79$ pairs of words are compared with the cosine distance of those word pairs based on different types of word embedding. The result shows that all the four types of word embeddings show a significant correlation with human affect judgment. The base BERT embedding has the weakest correlation of all, which also echoes with Fig.\ref{g:pca_4em}. Then, in the order of more relatedness is the GoEmotion BERT embedding, GloVe, and the contextualized BERT. It is worthwhile to deliberate on why the static word embeddings like GloVe outperforms the base BERT. Also, it is interesting that even though the GoEmotion BERT is fined tuned on emotion specific task, the correlation is still worse than BERT-based embeddings with movie specific contextualized information.

\begin{table*}[h]
\centering
\begin{tabular}{cccccc}
\hline
CORR (\emph{p}) & VAD         & GloVe       & base & GoEmotion & contextualized\\
\hline
VAD   & 1.000 (.000) &             &      & & \\
GloVe & \textbf{0.272} (.000) & 1.000 (.000) &      & &\\
base  & 0.116 (.000) & 0.148 (.000) & 1.000 (.000) &&\\
GoEmotion & \textbf{0.252} (.000) & 0.172 (.000) & 0.013 (.471) & 1.000 (.000) & \\
contextualized & \textbf{0.314} (.000) & \textbf{0.710} (.000) & \textbf{0.240} (.000) & \textbf{0.204} (.000) & 1.000 (.000)\\
\hline
\end{tabular}
\caption{Spearman Correlation (\textit{p} value) of pairwise cosine similarities between each and the rest of word embedding types (Coefficients larger than 0.2 and less than 1 are in bold.)}
\label{correlation}
\end{table*}

\subsection{Linear Classifier}

In this section, we report the classification results from the linear classifier probe. The training covariate matrix of the classification model was each type of word embeddings. The response variables were the binary categorical label representing lower or higher range of one of the VAD dimensions. The conversion was based on a 0.5 threshold. Each type of word embedding with each affect dimension constituted a linear classifier and together we constructed 12 classifiers. We used the NRC VAD vocabulary as the training and the validation data (with a 70/30 split). The test data comprised of 130 word with strong affect information randomly extracted from \cite{shaver1987emotion}.

We compared the performance of GloVe, base BERT, GoEmotion BERT, and contextualized BERT embeddings on each of the three affect dimensions and present the result in Table \ref{table:1}. It is clear that the contexualized BERT embedding shows superior performance compared to the others. To be noted, our validation set included many words whose VAD score might not be as salient and intuitive as the ones in the test set. This might explain that the validation set received lower accuracy score than the test sample. 

\begin{table*}
\centering
\begin{tabular}[ht]{c|c|c|c|c|c|c}
\hline
 \multicolumn{1}{c|}{Embedding types} &\multicolumn{3}{|c|}{Validation accuracy}
&\multicolumn{3}{|c}{Affect word sample accuracy}
\\
\hline
 & Valence & Arousal & Dominance & Valence & Arousal & Dominance   \\ 
\hline
GloVe  & 0.75 &	0.70	& 0.73 & 0.84 &	0.74 & 0.75 \\
\hline
base BERT &  0.76 &	0.73 &	0.74 &	0.93 &	0.85 &	0.82  \\  
\hline
GoEmotion BERT & 0.68 &	0.68 &	0.70 &	0.92 &	0.76 &	0.76 \\
\hline
contextualized BERT &  \textbf{0.85}& \textbf{0.77}&	\textbf{0.85}&	\textbf{0.95}&	\textbf{0.88}&\textbf{0.90}\\
\hline
\end{tabular}
\caption{Performance of different word embeddings in predicting VAD labels.}
\label{table:1}
\end{table*}

Besides, we observed that all of the model embeddings are best at predicting the valence dimension compared with arousal and dominance. This matches our prior knowledge because valence represents the positive or negative of the words and carries the most straightforward meaning for the context to capture. But arousal and dominance will be more complicated and subtle to capture depending on its distributional semantics. Interestingly, GoEmotion BERT does not perform very well in the dimensions of arousal and dominance. This phenomenon leaves ample space to investigate further how word embeddings are trained to capture certain meaning but not the others.

\section{Discussion \& Conclusion} \label{discussion}

In this work, we investigated the capability of word embeddings to capture affect meaning. By combining results from unsupervised and supervised learning, we found that our contextualized BERT embedding, where the representation of each word is the first principal component of all embeddings of the same word in its multiple occurrences in the IMDB dataset. In comparison, the base BERT does not show prominent encoding of affect meaning and sometimes is even inferior to the static GloVe. The GoEmotion BERT embedding also does not perform as well as the contextualized BERT embedding. Out of the VAD dimensions, it is hard to know which dimension is relatively well represented or easy to capture: based on the PCA results, all VAD dimensions could attain a correlation higher than 0.2 in some embedding representations; based on the linear classifier results, valence seems to be better represented than the other two. The discrepant pattern might result from the different sample sizes used in these tasks (i.e., 5,586 words for PCA and 130 words as the test set for the linear classifiers) and more controlled experiments are needed to pin down the fact. 
\subsection{Implications}
Here we provide a novel aspect of testing word embedding capabilities that involve intricate emotion and affect related meaning. Based on what lexical and linguistic knowledge word embeddings can and cannot capture, especially for the renowned BERT embedding \citep{ettinger2020bert, klafka2020spying, manning2020emergent,pandia2021pragmatic}, we add one additional piece of information. The relatively poor encoding of affect information in BERT might suggest these lines of implications: (1) the affect information of English lexicon is by nature implicit and hard to capture from the distributional features of words in huge corpora; (2) for sentiment related NLP tasks or affect computing tasks \citep{maas-EtAl:2011:ACL-HLT2011, socher2013recursive, zhang2015character}, we might consider using affect enriched embeddings or leveraging additional emotional capabilities of LMs from transfer learning.
\subsection{Limitations \& Future work}
As far as we know, this is the first attempt in the NLP community to investigate the capability of word embeddings to encode affect information in words. We lay three lines of research for future investigations. First, from an algorithmic point of view, we would like to ask what specific training mechanisms enable word embeddings to capture certain information, especially those intricate meanings that don't rely on co-occurrence and distributional semantics. Second, talking about meaning and semantics in a broader sense, there are still tons of unknowns that fall outside of the realm of formal semantics or compositionality. More rigorous identifications and descriptions of different kinds of meaning are needed for studying the knowledge of meaning in LMs. Third, to what extent can language carry affective information? This is an essential question in affective computing \citep{picard2000affective, picard2003affective, tao2005affective}. A better understanding from both the algorithmic level and the linguistic level of affect encoding and recognition would provide valuable insights into this question.
 
\setlength{\bibhang}{0.3in}			
\titleformat{\section}{\normalfont\bfseries}{\thesection}{.5em}{}		

\bibliographystyle{sp}		
\newcommand{\doi}[1]{\href{#1}{#1}}	
\vspace{6pt}

\bibliography{main_v2} 

\end{document}

%% file: Preamble.tex
\usepackage{leipzig}
\usepackage{tipa} 
\usepackage{graphicx}
\usepackage{linguex}
\usepackage{multirow}

\usepackage{enumitem}
\setlist{noitemsep, topsep=6pt} 

\usepackage{times}
\usepackage{natbib}	
\setcitestyle{comma,aysep={},yysep={,},notesep={; }}
\usepackage{fullpage} 
\usepackage[compact]{titlesec}
	\titleformat{\section}[runin]{\normalfont\bfseries}{\thesection.}{.5em}{}[.]
	\titleformat{\subsection}[runin]{\normalfont\scshape}{\thesubsection.}{.5em}{}[.]
	\titleformat{\subsubsection}[runin]{\normalfont\scshape}{\thesubsubsection.}{.5em}{}[.]
\usepackage[usenames,dvipsnames]{color}	
\definecolor{splinkcolor}{rgb}{.0,.2,.4}
\usepackage[colorlinks,allcolors={black},urlcolor={splinkcolor}]{hyperref} 		

\usepackage[vskip=6pt,leftmargin=0.3in,rightmargin=0.3in,indentfirst=false]{quoting}

\makeatletter
\def\@maketitle{%
  \newpage
  \begin{center}%
  \let \footnote \thanks
    {\normalfont\bfseries \@title \par}%
    {\vskip .5em
      \begin{tabular}[t]{c}%
        \@author
      \end{tabular}\par}%
  \end{center}%
  \par}
\makeatother
\renewenvironment{abstract}{%
\noindent\begin{minipage}{1\textwidth}
\setlength{\leftskip}{0.4in}
\setlength{\rightskip}{0.4in}
\textbf{Abstract.}}
{\end{minipage}}
\newenvironment{keywords}{%
\vspace{.5em}
\noindent\begin{minipage}{1\textwidth}
\setlength{\leftskip}{0.4in}
\setlength{\rightskip}{0.4in}
\textbf{Keywords.}}
{\end{minipage}}